\documentclass[a4paper,11pt]{article}

\usepackage{graphicx}  
\usepackage{url}       
\usepackage{times}
\usepackage{natbib}
\usepackage[margin=25mm]{geometry}
\usepackage{latexsym}
\usepackage{amssymb}
\usepackage{amsmath}
\usepackage[framemethod=TikZ]{mdframed}
\usepackage{mathtools}
\usepackage{floatrow}
\usepackage{subcaption}
\usepackage{wrapfig}
\usepackage{todonotes}
\usepackage{esvect}
\usepackage{multirow}
\usepackage{hhline}
\usepackage{pifont}
\usepackage{tikz}
\usepackage{pgfplots}
\usepackage{colortbl}
\usepackage{pgfplotstable}
\usepackage{filecontents}
\usetikzlibrary{pgfplots.groupplots}
\pgfplotsset{compat=1.3}
\usepackage{graphics}
\usetikzlibrary{intersections,positioning}
\usepackage{tikz-dependency}
\usepackage{color}
\usepackage{booktabs,amsfonts,dcolumn}
\definecolor{gray1}{gray}{0.8}
\definecolor{gray2}{gray}{0.5}
\definecolor{gray3}{gray}{0.1}
\newcommand*{\affaddr}[1]{#1} 
\newcommand*{\affmark}[1][*]{\textsuperscript{#1}}
\newcommand*{\email}[1]{\texttt{#1}}

\title{Cross-Lingual Transfer of Semantic Roles: From Raw Text to Semantic Roles }

\author{%
Maryam Aminian\affmark[1], Mohammad Sadegh Rasooli\affmark[2], Mona Diab\affmark[1]\\
\affaddr{\affmark[1]Department of Computer Science, The George Washington University, Washington}\\
\affaddr{\affmark[2]Facebook AI, Menlo Park, CA}\\
\email{\affmark[1]\{aminian,mtdiab\}@gwu.edu},\email{\affmark[2]rasooli@fb.com}\\
}

\date{}

\begin{document}
\maketitle
\begin{abstract}
We describe a transfer method based on annotation projection to develop a dependency-based semantic role labeling system for languages for which no supervised linguistic information other than parallel data is available. Unlike previous work that presumes the availability of supervised features such as lemmas, part-of-speech tags, and dependency parse trees, we only make use of word and character features. Our deep model considers using character-based representations as well as unsupervised stem embeddings to alleviate the need for supervised features. Our experiments outperform a state-of-the-art method that uses supervised lexico-syntactic features on 6 out of 7 languages in the Universal Proposition Bank.
\end{abstract}

\section{Introduction}
Despite considerable efforts on developing semantically annotated resources for semantic role labeling (SRL)~\citep{palmer_propbank,erk_german_srl,zaghouani_arabic}, majority of languages do not have such annotated resources. The lack of annotated resources for SRL has led to a growing interest in transfer methods for developing semantic role labeling systems. The ultimate goal of transfer methods is to transfer supervised linguistic information from a rich-resource language to a target language of interest. Amongst transfer methods, annotation projection is a method that projects supervised annotation from a rich-resource language to a low-resource language through automatic word alignments in parallel data~\citep{hwa2002evaluating,pado2009cross}. Recent work on annotation projection for SRL~\citep{kozhev_bootstrap,vanderplas_cross_global,akbik_projection,aminian-rasooli-diab:2017:I17-2} presumes the availability of accurate supervised features such as lemmas, part-of-speech (POS) tags and syntactic parse trees. However, this is not a realistic assumption for truly low-resource languages, for which (accurate) supervised features are hardly available.

This paper considers the problem of annotation projection of {\emph{dependency-based}} SRL in a scenario for which {\emph{only}} parallel data is available for the target language. Recent state-of-the-art SRL systems have shown a significant reliance on the predicate lemma information while in a low-resource language, a lemmatizer might not be available. We first demonstrate that unsupervised stems can be used as an alternative to supervised lemma features. We further show that we can obtain a robust and simple SRL model for the target language without relying on {\emph {any}} explicit linguistic feature (including lemmas), either supervised or unsupervised. We achieve this goal by changing the structure of a state-of-the-art deep SRL system~\citep{marcheggiani-2017} to make it independent of supervised features. Our model solely rely on word and character level features in the target language.

The main contribution of this work is on applying annotation projection without relying on supervised features in the target language of interest.  To the best of our knowledge, this is the first study that builds a cross-lingual SRL transfer model in the absence of any explicit linguistic information in the target language. We make use of the recently released Universal Proposition Banks~\citep{D16-1102}\footnote{\url{https://github.com/System-T/UniversalPropositions}}, a semi-automatically annotated data that unifies the annotation scheme for all languages.  We show the effectiveness of our method on a range of languages, namely German, Spanish, Finnish, French, Italian, Portuguese, and Chinese. We compare our model to a state-of-the-art baseline that uses a rich set of supervised features and show that our model outperforms on six out of seven languages in the Universal Proposition Banks. Furthermore, for Finnish, a morphologically rich language, our model with unsupervised features improves over the model that relies on a supervised lemmatizer.

This paper is structured as the following: \S\ref{sec:background} briefly overviews the dependency-based SRL task and annotation projection, \S\ref{sec:approach} describes our approach, \S\ref{sec:experiments} shows the experimental results and analysis, \S\ref{sec:related} gives overviews about the related work, and \S\ref{sec:conclusion} concludes the paper and proposes suggestions for future work.

\section{Background}\label{sec:background}
In this section, we provide a brief overview of dependency-based SRL and annotation projection. 

\paragraph{Dependency-based SRL} In dependency-based SRL, the goal is to find arguments along with their roles for each predicate in a sentence. Formally, in a sentence $x = [x_i]_{i=1}^{n}$ with $n$ words, and $m$ predicates $\mathbb{P} =[(p_i, \psi_i); 1 \leq p_i \leq n]_{i=1}^{m}$ where $\psi_i$ is the \emph{sense} of the predicate with index $p_i$ in the sentence, we find the semantic dependencies between each word in the sentence with respect to each predicate:
\[
\mathbb{L}_{x} = [ (p_i \xrightarrow{r} j | \psi_i); 1 \leq j \leq n, p_i \in \mathbb{P} ]
\]
where $r$ is the role of the $j$th word as an argument for the predicate word $x_{p_i}$. In case that a word is not an argument, $r$ is $\tt{NULL}$. Evaluation of the system output is conducted on semantic dependencies $(p_i \xrightarrow{r} j | \psi_i)$; thus the SRL system should find  predicate senses as well as argument roles. During training, these dependencies are used as training instances for a machine learning algorithm. Previous work~\citep{bj_sup,roth_sup,marcheggiani-2017} factorized this task into predicate sense disambiguation, argument identification, and argument classification.  

\begin{figure*}
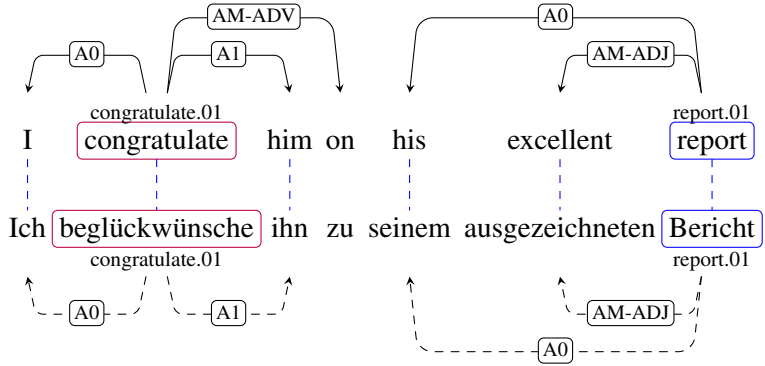


\scalebox{1}{
\centering
\begin{dependency}
\begin{deptext}
\&\scriptsize{congratulate.01}\&\&\&\&\&\scriptsize{report.01}\\
I\&congratulate\&him\&on\&his\&excellent\&report\\
\\
\\
Ich\&begl\"uckw\"unsche\&ihn\&zu\&seinem\&ausgezeichneten\&Bericht\\
\&\scriptsize{congratulate.01}\&\&\&\&\&\scriptsize{report.01}\\
\end{deptext}

\path[dashed,color=blue] (\wordref{2}{1}) edge (\wordref{5}{1});
\path[dashed,color=blue] (\wordref{2}{2}) edge (\wordref{5}{2});
\path[dashed,color=blue] (\wordref{2}{5}) edge (\wordref{5}{5});
\path[dashed,color=blue] (\wordref{2}{3}) edge (\wordref{5}{3});
\path[dashed,color=blue] (\wordref{2}{7}) edge (\wordref{5}{7});
\path[dashed,color=blue] (\wordref{2}{6}) edge (\wordref{5}{6});

\depedge{2}{1}{A0}
\depedge{2}{3}{A1}
\depedge{2}{4}{AM-ADV}
\depedge{7}{6}{AM-ADJ}
\depedge{7}{5}{A0}

\depedge[dashed,edge below]{2}{1}{A0}
\depedge[dashed,edge below]{2}{3}{A1}
\depedge[dashed,edge below]{7}{6}{AM-ADJ}
\depedge[dashed,edge below]{7}{5}{A0}

\wordgroup[color=purple]{2}{2}{2}{pred}
\wordgroup[color=blue]{2}{7}{7}{pred}

\wordgroup[color=purple]{5}{2}{2}{pred}
\wordgroup[color=blue]{5}{7}{7}{pred}

\end{dependency}
}
\caption{An example of annotation projection for an English-German sentence pair from the Europarl corpus~\protect\citep{koehn2005europarl}. Supervised predicate-argument structure of the English sentence (edges on top) is generated using our supervised SRL system trained on PropBank 3~\protect\citep{palmer_propbank}. Dashed lines in the middle show intersected word alignments from Giza++~\protect\citep{och2000giza}. Dashed edges at the bottom show the projected predicate-arguments.}
\label{fig:proj}
\end{figure*}

\paragraph{Annotation Projection} In annotation projection, we assume that we have a parallel data ${\cal P} = [ (s^{(1)}, t^{(1)}),\\ \cdots, (s^{(k)}, t^{(k)})]$ such that each sentence $s^{(i)}$ is a translation of sentence $t^{(i)}$. Here, we assume that  $s^{(i)}$ belongs to a rich-resource language in which annotated resources are available. In contrast, $t^{(i)}$ belongs to a low-resource target language where annotated data and tools such as semantic roles, dependency trees, part-of-speech tags, word senses, and lemmas  might not be available. 


For every sentence $s^{(i)}$, we run a supervised SRL system to obtain its supervised argument structure $\mathbb{L}_{s^{(i)}}$. Assuming that  $s^{(i)} = [s^{(i)}_1, \cdots, s^{(i)}_{l_i}]$ and  $t^{(i)} = [t^{(i)}_1, \cdots, t^{(i)}_{l'_i}]$, we use an automatic word alignment system to obtain \emph{one-to-one} word alignments. We define $0 \leq a^{(i)}_{j} \leq l_i$ as the index of the source word that is aligned to the $j$th word in the $i$th target sentence, where $a^{(i)}_{j}=0$  indicates a missing alignment.  We use the following conditions to project a semantic dependency from a source sentence to a target sentence:
\[
(a^{(i)}_p \xrightarrow{r} a^{(i)}_m | y) \in \mathbb{L}_{s^{(i)}}  \Rightarrow \text{add } (p \xrightarrow{r} m | y) \text{ to } \mathbb{L}_{t^{(i)}} 
\]
where $\mathbb{L}_{s^{(i)}}$ is the supervised argument structure and $\mathbb{L}_{t^{(i)}}$ is the projected argument structure for the $i$th sentence. We assume that there is a universal predicate sense that is common across languages (this is the case in the Universal Propositon Banks). Figure \ref{fig:proj} shows an example for an English-German translation pair. We use the projected data as training data in a supervised learning system to train a SRL system in the target language. In practice, many words do not receive any projected label mainly due to missing alignments. Thus, $\mathbb{L}_{t^{(i)}}$ usually contains sentences with partially projected semantic dependencies. 


\section{Our Model}\label{sec:approach}
Our goal is to train a SRL system on the projected predicate-argument structures without having supervised features such as supervised lemmas, dependency parse trees, and part-of-speech tags. Our model has two main components: 1) joint argument identification and classification which we simply refer to as argument classifier , and 2) predicate sense disambiguation. 
Our argument classifier is inspired by the model of \citet{marcheggiani-2017}: we use predicate-specific BiLSTM encoders, and a role+predicate-specific decoder. However, unlike the model of \citet{marcheggiani-2017}, which relies heavily on POS tags and predicate lemmas, we do not use a supervised lemmatizer and POS tagger in any layer. Instead, we benefit from character representations and unsupervised stems to bring in unsupervised features to our model. 
\subsection{Joint Argument Identification and Classification}
Given a sentence $s = [s_i]_{i=1}^{n}$ that contains $n$ tokens with $m$ predicates in the predicate set $\mathbb{P}$, we run $m$ \emph{separate} predicate-specific deep BiLSTM encoders $[\mathbb{E}_j]_{j=1}^{m}$ to extract contextualized representations for each token given a predicate index $p_j$. 

\paragraph{Input Representation}
For each encoder $[\mathbb{E}_j]_{j=1}^{m}$, we represent each token $s_i$ as the concatenation of its word embedding ($x_i^{re}$ and $x_i^{pe}$), character embedding ($x_i^{char}$) and predicate lemma embedding ($x_{i,j}^{lem}$):\footnote{We use [;] notation to show vector concatenation.}

\[
\begin{split}
    x_{i,j} =  [x_i^{re}; x_i^{pe}; x_i^{char}; x_{i,j}^{le}]& \\
    \forall i \in [1,\cdots,n]; & ~ j \in [1, \cdots, m]
\end{split}
\]
where:

\begin{itemize}
    \item $x_i^{re} \in \mathbb{R}^{d_w}$ is a randomly initialized word embedding vector;
    \item $x_i^{pe} \in \mathbb{R}^{d_w} $ is an external pre-trained word embedding that is fixed during training;
    \item $x_i^{char} \in \mathbb{R}^{d_{ch}}$ is character representation of each token $s_i$. For every token,
     we obtain $x_i^{char}$ by running a deep bidirectional LSTM~\citep{hochreiter1997long} on top of each word. We use the concatenation of the final backward representation of the first character, and final forward representation of the last character to represent each token:  
\[
x_i^{char} = {\tt BiLSTM} (x_i^c[1:|s_i|] ; |s_i|)
\]
where $x_i^{c}  \in \mathbb{R}^{d_c}$ is a randomly initialized character embedding  and $|s_i|$ is the number of characters in token $s_i$;
    \item $x_{i,j}^{le} \in \mathbb{R}^{d_{le}}$ is a lemma vector for each word $s_i$ with respect to the predicate that is targeted in $\mathbb{E}_j$. $x_{i,j}^{le}$ is active if $s_i$ is the predicate word, otherwise, a zero vector is used to represent the lemma embedding:

\[
x_{i,j}^{le} = 
\begin{cases}
[x_i^{le}; 1] & \tt{if}~ i = p_j \\
[\overrightarrow{0}; 0] & {\tt otherwise} \\ 
\end{cases}
\]

where the concatenated zero/one value is a flag to indicate if the current token is the targeted lemma. In our model, we use one of the following options to represent predicate lemma:

\begin{itemize}
    \item Represent each lemma by a deep character BiLSTM. This BiLSTM is different from the character BiLSTM in $x^{char}$. 
    \item Use an unsupervised morphological analyzer to give the surface-form stem of each word. This way, we can use a lemma embedding dictionary without requiring a lemmatizer.
\end{itemize}
    
\end{itemize}

\paragraph{Predicate-Specific Encoder}
A deep BiLSTM is used to get the final representation for each token in a sentence. In the following notation, $h_{i,j}$ is the final hidden state from the deep BiLSTM model for the $i$th token with respect to the $j$th predicate:

\[
h_{i,j} = {\tt BiLSTM}(s_{1:n, j}; i) \in \mathbb{R}^{d_h}
\]

\paragraph{Role+Predicate-Specific Decoder}

Given the BiLSTM representations, we perform an affine transformation on the concatenation of $h_{p_j, j}$ (predicate representation) and $h_{i, j}$ (argument representation) to find the probability of having the $i$th token as the argument of predicate $p_j$ with role $r$ (including the {\tt NULL} role):
\[
p(r | h_{p_j, j}, h_{i,j}) = {\tt softmax}_{\tt r}(W_{j,r} [h_{p_j, j}; h_{i,j}])
\]
where $x_{j,r} $ is  a parameter matrix that encodes the information of role $r$ and the $j$th predicate. This matrix is calculated as follows:
\[
W_{j,r} = {\tt RELU} (U [u^{l}_{j}, v_{r}])
\]
where $u^{l}_{j} \in \mathbb{R}^{d'_{l}}$ is another predicate lemma embedding parameter which is specifically used for the decoder layer, $v_{r} \in {\mathbb{R}^{d_r}}$ is a randomly initialized role embedding,  $U$ is a parameter matrix, and {\tt RELU} is the rectified linear units activation function~\citep{nair2010rectified}. Similar to the input layer, we represent $u^{l}_{j}$ by 1) a different deep character BiLSTM, \emph{or} 2) a surface-form stem obtained from an unsupervised morphological analyzer.





\begin{figure*}[t!]
\floatbox[{\capbeside\thisfloatsetup{capbesideposition={left,top},capbesidewidth=5cm}}]{figure}[\FBwidth]
{\caption{A graphical depiction of our joint argument identification and classification model without using part-of-speech tags, lemmas, and syntax. In this example, the predicate-specific encoder considers word \emph{eats} as the sentence predicate and the goal is to score the assignment of argument \emph{apple} with label $A_0$. Our model contains three \emph{different} character BiLSTMs; at the bottom, a character BiLSTM is run to acquire a character-based representation for all the words in the sentence in the absence of POS tags. There are two character BiLSTMs for predicate lemma: one in the encoder level (next to the second word) to model predicate lemma in the input layer and the other in the decoder level (top left). In this example, we just show one layer of BiLSTM but we use a deep BiLSTM in our experiments.}\label{fig_network}}
{   \includegraphics[height=0.6\textheight]{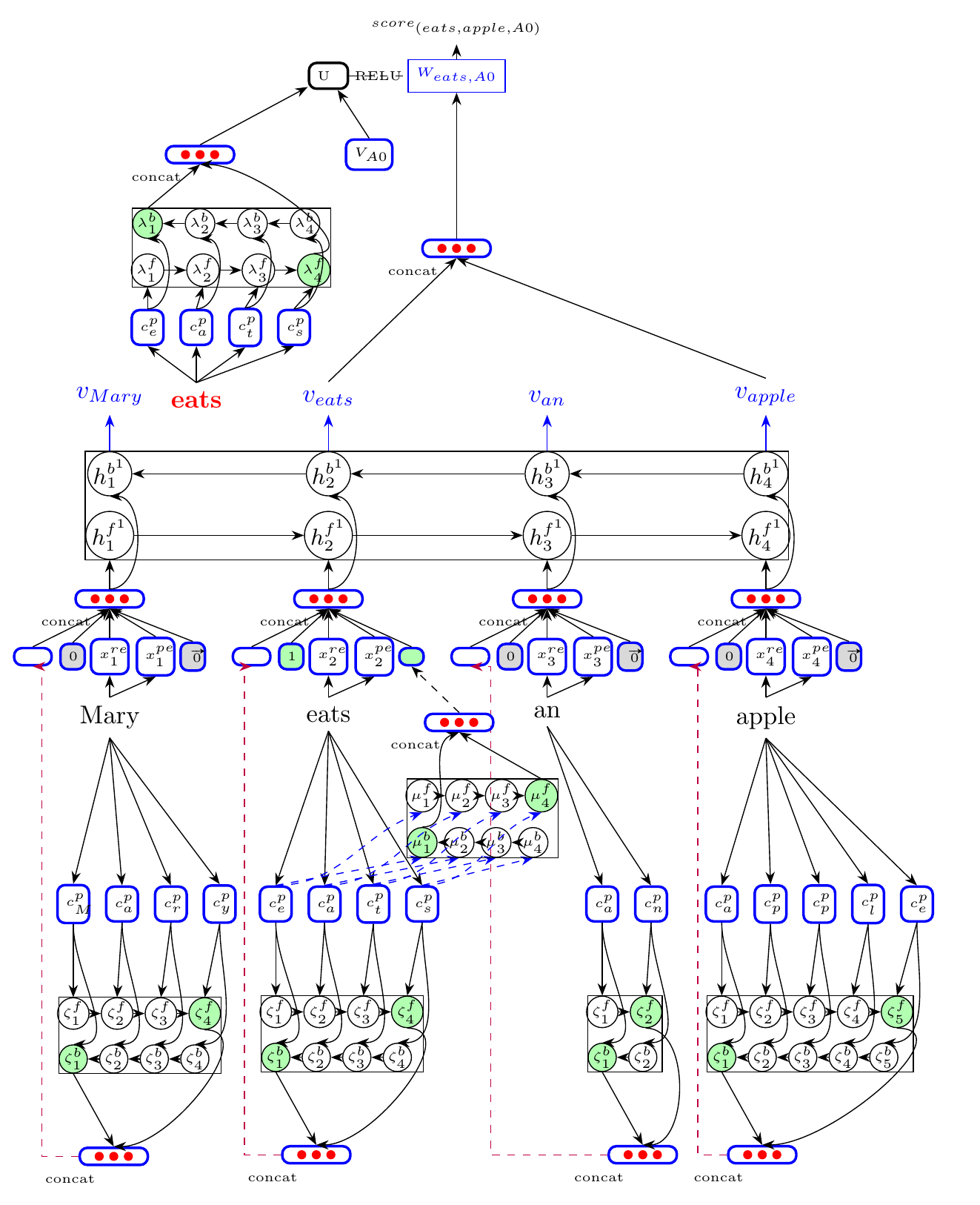}}
\end{figure*}

A graphical depiction of the network in a case for which lemmas are represented by character BiLSTMs is shown in figure~\ref{fig_network}. As shown in the figure, we use two different character BiLSTMs in order to represent lemmas: one for the input representation and the other for the decoder representation.  



\section{Experiments}\label{sec:experiments}
\paragraph{Datasets and Tools}
We use English as the source language and project SRL annotations to the following languages: German, Spanish, Finnish, French, Italian, Portuguese, and Chinese. We use the Europarl parallel corpus \citep{koehn2005europarl} for the European languages and a random sample of 2 million sentence pairs from the MultiUN corpus~\citep{multiun} for Chinese. We use the Giza++ tool~\citep{och2000giza} with its default setting for word alignment. We run Giza++ in source-to-target and the reverse direction and get the intersection of alignment links. 
For English, we use the pre-trained embedding vectors generated using the structured skip-gram model of \citet{ling-EtAl:2015:NAACL-HLT}. For the target languages, we train Word2vec~\citep{mikolov2013distributed} on Wikipedia data to generate embedding vectors.

We implement our deep network using the Dynet library~\citep{neubig2017dynet}. We use the dimension of 100 for word embeddings, 50 for characters, 512 for LSTM encoders, 128 for role and lemma embeddings in the decoder, and 100 for decoder lemma embedding. We pick random minibatches of size 1000 with a fixed learning rate of $0.001$ for learning the parameter values with the Adam optimizer~\citep{kingma2014adam}. The depth of BiLSTM network is set to one for character representation ($x^{char}$) and three for predicate-specific representations ($x^{le}$, $u^l$).

\paragraph{Predicate Disambiguation}\label{sec:pdism}
Our model is agnostic to predicate senses but since our automatic evaluation relies on automatic predicate senses, we need a disambiguation module. Predicate disambiguation systems typically contains separate classifiers for each predicate lemma~\citep{bj_sup}. Since we do not have a reliable lemmatizer in the target language, we train a single classifier for all predicates. We encode a sentence with a three-layer deep BiLSTM and run a softmax layer on top of each predicate to disambiguate the predicate sense of each predicate.

\paragraph{Predicate identification on the source side} For projection experiments, first of all we need to identify predicates in the source language. Input to our predicate identifier is the concatenation of word embedding, pre-trained fixed word embedding, POS embedding\footnote{Since this is only used for a supervised setting, we are able to use POS features.}, and character representation (obtained from a character BiLSTM) for every token in the sentence. We use a deep BiLSTM to get the final representation for each token. The ultimate predictions are made by performing an affine transform on the BiLSTM hidden output.

\subsection{Projection Experiments} 
Our supervised SRL system is a reimplementation of the model of \citet{marcheggiani-2017}. We generate automatic English predicate senses using a system similar to the predicate disambiguation module of \citet{bj_sup} except that we replace the logistic regression classifier with the averaged Perceptron algorithm \citep{collins:2002:EMNLP02}. In order to comply with the Universal Proposition Bank annotation scheme, we convert the argument spans in the English PropBank v3 \citep{palmer_propbank} to dependency-based arguments by labeling the syntactic head of each span.

For annotation projection, we define density of alignments to find sentences with relatively-dense alignments:
\[
 {\tt density}^{(i)}= \frac{\sum_{j=1}^{l'_j} \mathbb{I}(a_j^{(i)}>0) }{l'_i}
 \]
where $l'_i$ is the length of the $i$th target sentence in parallel data, $a_j^{(i)}$ is the alignment index for the $j$th word in the target sentence, and $\mathbb{I}(a_j^{(i)}>0)$ is an indicator for a non-{\tt NULL} alignment. We prune the target sentence pairs with density less than 80\% for all European languages. We set this threshold to 60\% for Chinese in order to obtain a comparable number of sentences to the European languages. Table~\ref{table:proj:stat} summarizes the sizes of projected datasets after applying the density filter. We set the number of training epochs to 2 for all languages based on development results obtained from the English to German projections.

Since the original model of \citet{marcheggiani-2017} heavily relies on the predicate lemma information for making robust prediction, we further assess the influence of using explicit linguistic features in our model by using a) supervised lemma from the UDPipe pre-trained models~\citep{udpipe:2017}, and b) unsupervised stems obtained from unsupervised morphological analyzer. We use the unsupervised morphological analyzer of \citet{virpioja2013morfessor}, and obtain morpheme classes by running Morfessor FlatCat~\citep{gronroos-EtAl:2014:Coling} on the output of the analyzer. We run the \emph{fixed-affix} finite-state machine of \citep{rasooli-EtAl:2014:P14-1} to obtain a single stem for all words including the out-of-vocabularies.

\begin{table}[t!]
\centering
\label{table:proj_stat}
\small
\begin{tabular}{l | llll}
\hline \hline
Lang. & \#\footnotesize{Sent.} & \#\footnotesize{Tokens} & \#\footnotesize{Types} & \#\footnotesize{Pred.}\\
\hline
de & 332K  & 6M  & 90K & 867K \\
es & 903K  & 25M & 120K & 3M \\
fi & 558K & 8M & 243K & 1M \\
fr & 924K & 26M & 93K & 3M \\
it & 617K & 17M  & 88K & 2M \\
pt & 632K & 17M & 98K & 2M \\
zh & 821K & 21M & 183K  & 1M \\
\hline \hline
\end{tabular}
\caption{Sizes of the projected data.}\label{table:proj:stat}
\end{table}

\begin{table*}[t!]
\centering
\label{table:proj}
\small
\setlength{\tabcolsep}{8pt}
\begin{tabular}{l c c c cccc}
\hline\hline

System & de &  es & fi & fr & it & pt & zh\\
\hline

Bootstrap &
59.8 \scriptsize{(55.0)} & 60.6 \scriptsize{(52.2)} & 59.0 \scriptsize{(53.1)} & 71.0 \scriptsize{(63.4)} &  59.2 \scriptsize{(52.3)} & 61.2 \scriptsize{(53.9)} & 50.3 \scriptsize{(42.5)}\\
\hline
SLem & 61.7 \scriptsize{(57.0)} & 62.4 \scriptsize{(55.7)} & 62.5 \scriptsize{(59.2)} & 65.0 \scriptsize{(58.9)} & 61.8 \scriptsize{(56.4)} & 63.0 \scriptsize{(56.8)}  & 52.1 \scriptsize{(43.7)}\\
UStem & 62.0 \scriptsize{(57.4)} & 63.0 \scriptsize{(56.0)} & 64.5 \scriptsize{(58.8)} & 65.3 \scriptsize{(59.2)} & 61.3 \scriptsize{(55.4)} & 62.8 \scriptsize{(56.8)} & 52.6 \scriptsize{(43.2)}\\
\hline
{\bf CModel} & 61.0 \scriptsize{(57.0)} & 62.5 \scriptsize{(56.0)} & 64.6 \scriptsize{(58.9)} & 65.1 \scriptsize{(58.5)} & 61.0 \scriptsize{(55.5)} & 62.9 \scriptsize{(56.5)} & 52.7 \scriptsize{(42.7)}\\
\hline
Supervised & 74.5 \scriptsize{(72.0)} & 77.8 \scriptsize{(75.2)} & 74.0 \scriptsize{(69.6)} & 88.9 \scriptsize{(87.5)} &77.9 \scriptsize{(75.9)}& 66.6 \scriptsize{(62.4)} & 68.8 \scriptsize{(68.6)} \\

\hline\hline

\end{tabular}

\caption{Results of projection experiments using our character based model (\emph{CModel}) on the Universal PropBank test sets compared to different baselines: the SRL system of {\emph{\citet{aminian-rasooli-diab:2017:I17-2}}} ({\emph{Bootstrap}}), {\emph{SLem}} that shows the results of our model when supervised lemma is used and {\emph{UStem}} that show the results of our model with unsupervised stem. Numbers in parenthesis show results with automatic predicate senses.}\label{table:proj:results}   

\end{table*}

\paragraph{Results} 
We compare our character-based approach ({\emph{CModel}}) with three different models: 1) The cross-lingual model of \citet{aminian-rasooli-diab:2017:I17-2} ({\emph {Bootstrap}}) that uses a rich set of supervised features including supervised lemmas, POS tags, and dependency parse information, 2) a variant of our model that uses supervised lemmas ({\emph {SLem}}) generated by a lemmatizer to represent predicate lemmas in the input and the decode layers, and 3) a model similar to the second model but using unsupervised stems ({\emph {UStem}}) generated by an unsupervised morphological analyzer to represent predicate lemmas. Here, we aim to asses the effects of using different levels of explicit linguistic features ranging from fully specified supervised features to unsupervised features in our model. The {\emph {Bootstrap}} model uses an iterative bootstrapping approach by utilizing a special cost function and benefiting from a rich set of supervised lexico-syntactic features, thereby, it is considered a hard baseline. Since {\emph {Bootstrap}} has a large number of features, the model is not memory-wise scalable to our projection data sizes. Therefore we train the {\emph {Bootstrap}} model on a random sample of 20K sentences. This number is similar to the number of sentences used in the original experiments~\citep{aminian-rasooli-diab:2017:I17-2}.

Table \ref{table:proj:results} shows labeled F-scores using both gold and automatic predicate senses on the test portion of the Universal Proposition Banks. As shown in Table \ref{table:proj:results}, our model ({\emph{CModel}}) outperforms the {\emph{Bootstrap}} model for all languages except French. Additionally, our model performs on par to the supervised lemma and unsupervised stem models. This demonstrates the power of our approach even though our model has access to fewer linguistic features in the target language. Using unsupervised stems outperforms supervised lemma on all languages except Portuguese and Italian. This further highlights the reliance of the model on the accuracy of lemmatizer.

\begin{table*}[t!]
\centering
\begin{tabular}{llllllll}
Language & de & es & fi & fr & it & pt & zh \\
\hline\hline
\#Sentences & 14K & 14K & 12K & 14K & 12K & 938 & 3K \\
\#Tokens & 269K & 382K & 162K  & 356K & 270K  & 22K  & 98K \\
\#Types &49K & 46K & 49K & 42K & 28K & 6K & 17K \\
\#Predicates & 21K & 34K &25K & 29K &25K  & 2K & 10K \\
\hline
\end{tabular}
\caption{Sizes of treebanks in the Universal Propbank training data for all languages.}\label{table:uni_stats}
\end{table*}
\paragraph{Analysis} As shown in Table \ref{table:proj:results}, using automatic predicate senses leads to a significant reduction in accuracy. This degradation is caused by two reasons. First, training a single classifier for all predicates in the absence of explicit predicate lemma information, and second, using unified predicate senses for all languages leads to lower precision for out-of-vocabulary words. This happens due to the fact that we cannot make use of the default sense of predicate ({\tt lemma.01}). Among all the languages in our experiments, French is the only language that our model underperforms the \emph{Bootstrap} model. Our analysis on French shows that our model has not been able to correctly predict A0 and A1 arguments in 20\% and 30\% of cases, and labeled them as {\tt NULL}. 

\section{Related Work}\label{sec:related}
There has been a great deal of interest in using transfer methods for SRL by different techniques such as enhancing the quality of projections \citep{pado_projection, pado2009cross}, joint learning of syntax and semantics~\citep{vanderplas_cross,kozhevnikov_transfer}, and iterative bootstrapping to learn a robust model from erroneous projections~\citep{akbik_projection, aminian-rasooli-diab:2017:I17-2}.
 Previous work presumes availability of a wide range of supervised lexico-syntactic features for the target language. Consequently, their performance heavily relies on accuracy of the available tagging tools~\citep{akbik_projection}. For instance, \citet{akbik_projection} reports lower argument precision for languages that do not have accurate syntactic parsers such as Arabic and Hindi. In contrary to the previous studies, our work builds a cross-lingual SRL system without having any supervised features for the target language. 
 

One obstacle for developing transfer models is the absence of a unified annotation scheme for all languages. There has been a great deal of work in developing universal annotation schemes for a variety of tasks such as POS tagging~\citep{petrov2011universal}, dependency parsing~\citep{universal_2}, morphology~\citep{kirov2018unimorph}, and SRL~\citep{kozhev_bootstrap,wang-EtAl:2017:EMNLP20175}. Our work makes use of the recently released Universal Proposition Bank~\citep{D16-1102}. This dataset maps every predicate lemma in every language to its corresponding English lemma following the frame and role label schemes of the English Proposition Bank 3.0~\citep{palmer_propbank}

In the realm of \emph{supervised} SRL methods, however, there have been several efforts to build SRL models that do not need a wide range of linguistic features (specifically syntactic features) \citep{marcheggiani-2017,P15-1109,P17-1044,P18-1192,C18-1233,P18-2106}.  In a more recent study, \citet{P18-2106} proposed a polyglot SRL system that benefits from the similarities between the semantic structures of different languages to improve monolingual SRL. All those studies, however, assume the availability of semantically annotated datasets for the target language, thus making them non-applicable to low-resource languages.

\section{Conclusion}\label{sec:conclusion}
We have described a method for cross-lingual transfer of dependency-based SRL systems via annotation projection. Our model is agnostic to linguistic features leading to a robust model that can be trained on projected text on a target language without annotated data. We have shown that our model achieves comparable performance in annotation projection and also supervised SRL. In addition to improving the performance of our model with the current setting, future work should study more effective ways to apply the transfer methods; e.g. combining with the direct transfer method in the absence of large parallel corpora.

\section*{Acknowledgments}
The first and third authors have been partly funded by the DARPA LORELEI grant and generous support by Leidos Corp.. We would like to acknowledge the useful comments by three anonymous reviewers who helped in making this publication more concise and better presented.

\bibliographystyle{chicago}
\bibliography{ref}

\begin{thebibliography}{}

\bibitem[\protect\citeauthoryear{Akbik, chiticariu, Danilevsky, Li,
  Vaithyanathan, and Zhu}{Akbik et~al.}{2015}]{akbik_projection}
Akbik, A., l.~chiticariu, M.~Danilevsky, Y.~Li, S.~Vaithyanathan, and H.~Zhu
  (2015).
\newblock Generating high quality proposition banks for multilingual semantic
  role labeling.
\newblock In {\em Proceedings of the 53rd Annual Meeting of the Association for
  Computational Linguistics and the 7th International Joint Conference on
  Natural Language Processing (Volume 1: Long Papers)}, pp.\  397--407.
  Association for Computational Linguistics.

\bibitem[\protect\citeauthoryear{Akbik, kumar, and Li}{Akbik
  et~al.}{2016}]{D16-1102}
Akbik, A., v.~kumar, and Y.~Li (2016).
\newblock Towards semi-automatic generation of proposition banks for
  low-resource languages.
\newblock In {\em Proceedings of the 2016 Conference on Empirical Methods in
  Natural Language Processing}, pp.\  993--998. Association for Computational
  Linguistics.

\bibitem[\protect\citeauthoryear{Aminian, Rasooli, and Diab}{Aminian
  et~al.}{2017}]{aminian-rasooli-diab:2017:I17-2}
Aminian, M., M.~S. Rasooli, and M.~Diab (2017, November).
\newblock Transferring semantic roles using translation and syntactic
  information.
\newblock In {\em Proceedings of the Eighth International Joint Conference on
  Natural Language Processing (Volume 2: Short Papers)}, Taipei, Taiwan, pp.\
  13--19. Asian Federation of Natural Language Processing.

\bibitem[\protect\citeauthoryear{Bj{\"o}rkelund, Hafdell, and
  Nugues}{Bj{\"o}rkelund et~al.}{2009}]{bj_sup}
Bj{\"o}rkelund, A., L.~Hafdell, and P.~Nugues (2009).
\newblock {\em Proceedings of the Thirteenth Conference on Computational
  Natural Language Learning (CoNLL 2009): Shared Task}, Chapter Multilingual
  Semantic Role Labeling, pp.\  43--48.
\newblock Association for Computational Linguistics.

\bibitem[\protect\citeauthoryear{Cai, He, Li, and Zhao}{Cai
  et~al.}{2018}]{C18-1233}
Cai, J., S.~He, Z.~Li, and H.~Zhao (2018).
\newblock A full end-to-end semantic role labeler, syntactic-agnostic over
  syntactic-aware?
\newblock In {\em Proceedings of the 27th International Conference on
  Computational Linguistics}, pp.\  2753--2765. Association for Computational
  Linguistics.

\bibitem[\protect\citeauthoryear{Collins}{Collins}{2002}]{collins:2002:EMNLP02}
Collins, M. (2002, July).
\newblock Discriminative training methods for hidden {Markov} models: Theory
  and experiments with perceptron algorithms.
\newblock In {\em Proceedings of the 2002 Conference on Empirical Methods in
  Natural Language Processing}, pp.\  1--8. Association for Computational
  Linguistics.

\bibitem[\protect\citeauthoryear{Eisele and Chen}{Eisele and
  Chen}{2010}]{multiun}
Eisele, A. and Y.~Chen (2010, may).
\newblock Multiun: A multilingual corpus from united nation documents.
\newblock In N.~C.~C. Chair), K.~Choukri, B.~Maegaard, J.~Mariani, J.~Odijk,
  S.~Piperidis, M.~Rosner, and D.~Tapias (Eds.), {\em Proceedings of the
  Seventh conference on International Language Resources and Evaluation
  (LREC'10)}, Valletta, Malta. European Language Resources Association (ELRA).

\bibitem[\protect\citeauthoryear{Erk, Kowalski, Pad\'{o}, and Pinkal}{Erk
  et~al.}{2003}]{erk_german_srl}
Erk, K., A.~Kowalski, S.~Pad\'{o}, and M.~Pinkal (2003, July).
\newblock Towards a resource for lexical semantics: A large german corpus with
  extensive semantic annotation.
\newblock In {\em Proceedings of the 41st Annual Meeting of the Association for
  Computational Linguistics}, Sapporo, Japan, pp.\  537--544. Association for
  Computational Linguistics.

\bibitem[\protect\citeauthoryear{Gr\"{o}nroos, Virpioja, Smit, and
  Kurimo}{Gr\"{o}nroos et~al.}{2014}]{gronroos-EtAl:2014:Coling}
Gr\"{o}nroos, S.-A., S.~Virpioja, P.~Smit, and M.~Kurimo (2014, August).
\newblock Morfessor flatcat: An hmm-based method for unsupervised and
  semi-supervised learning of morphology.
\newblock In {\em Proceedings of COLING 2014, the 25th International Conference
  on Computational Linguistics: Technical Papers}, Dublin, Ireland, pp.\
  1177--1185. Dublin City University and Association for Computational
  Linguistics.

\bibitem[\protect\citeauthoryear{He, Lee, Lewis, and Zettlemoyer}{He
  et~al.}{2017}]{P17-1044}
He, L., K.~Lee, M.~Lewis, and L.~Zettlemoyer (2017).
\newblock Deep semantic role labeling: What works and what's next.
\newblock In {\em Proceedings of the 55th Annual Meeting of the Association for
  Computational Linguistics (Volume 1: Long Papers)}, pp.\  473--483.
  Association for Computational Linguistics.

\bibitem[\protect\citeauthoryear{He, Li, Zhao, and Bai}{He
  et~al.}{2018}]{P18-1192}
He, S., Z.~Li, H.~Zhao, and H.~Bai (2018).
\newblock Syntax for semantic role labeling, to be, or not to be.
\newblock In {\em Proceedings of the 56th Annual Meeting of the Association for
  Computational Linguistics (Volume 1: Long Papers)}, pp.\  2061--2071.
  Association for Computational Linguistics.

\bibitem[\protect\citeauthoryear{Hochreiter and Schmidhuber}{Hochreiter and
  Schmidhuber}{1997}]{hochreiter1997long}
Hochreiter, S. and J.~Schmidhuber (1997).
\newblock Long short-term memory.
\newblock {\em Neural computation\/}~{\em 9\/}(8), 1735--1780.

\bibitem[\protect\citeauthoryear{Hwa, Resnik, Weinberg, and Kolak}{Hwa
  et~al.}{2002}]{hwa2002evaluating}
Hwa, R., P.~Resnik, A.~Weinberg, and O.~Kolak (2002).
\newblock Evaluating translational correspondence using annotation projection.
\newblock In {\em Proceedings of the 40th Annual Meeting on Association for
  Computational Linguistics}, pp.\  392--399. Association for Computational
  Linguistics.

\bibitem[\protect\citeauthoryear{Kingma and Ba}{Kingma and
  Ba}{2014}]{kingma2014adam}
Kingma, D.~P. and J.~Ba (2014).
\newblock Adam: A method for stochastic optimization.
\newblock {\em arXiv preprint arXiv:1412.6980\/}.

\bibitem[\protect\citeauthoryear{Kirov, Cotterell, Sylak-Glassman, Walther,
  Vylomova, Xia, Faruqui, Mielke, McCarthy, K{\"u}bler, et~al.}{Kirov
  et~al.}{2018}]{kirov2018unimorph}
Kirov, C., R.~Cotterell, J.~Sylak-Glassman, G.~Walther, E.~Vylomova, P.~Xia,
  M.~Faruqui, S.~Mielke, A.~D. McCarthy, S.~K{\"u}bler, et~al. (2018).
\newblock Unimorph 2.0: Universal morphology.
\newblock {\em arXiv preprint arXiv:1810.11101\/}.

\bibitem[\protect\citeauthoryear{Koehn}{Koehn}{2005}]{koehn2005europarl}
Koehn, P. (2005).
\newblock Europarl: A parallel corpus for statistical machine translation.
\newblock In {\em MT summit}, Volume~5, pp.\  79--86.

\bibitem[\protect\citeauthoryear{Kozhevnikov and Titov}{Kozhevnikov and
  Titov}{2013a}]{kozhev_bootstrap}
Kozhevnikov, M. and I.~Titov (2013a).
\newblock Bootstrapping semantic role labelers from parallel data.
\newblock In {\em Second Joint Conference on Lexical and Computational
  Semantics (*SEM), Volume 1: Proceedings of the Main Conference and the Shared
  Task: Semantic Textual Similarity}, pp.\  317--327. Association for
  Computational Linguistics.

\bibitem[\protect\citeauthoryear{Kozhevnikov and Titov}{Kozhevnikov and
  Titov}{2013b}]{kozhevnikov_transfer}
Kozhevnikov, M. and I.~Titov (2013b).
\newblock Cross-lingual transfer of semantic role labeling models.
\newblock In {\em Proceedings of the 51st Annual Meeting of the Association for
  Computational Linguistics (Volume 1: Long Papers)}, pp.\  1190--1200.
  Association for Computational Linguistics.

\bibitem[\protect\citeauthoryear{Ling, Dyer, Black, and Trancoso}{Ling
  et~al.}{2015}]{ling-EtAl:2015:NAACL-HLT}
Ling, W., C.~Dyer, A.~W. Black, and I.~Trancoso (2015, May--June).
\newblock Two/too simple adaptations of word2vec for syntax problems.
\newblock In {\em Proceedings of the 2015 Conference of the North American
  Chapter of the Association for Computational Linguistics: Human Language
  Technologies}, Denver, Colorado, pp.\  1299--1304. Association for
  Computational Linguistics.

\bibitem[\protect\citeauthoryear{Marcheggiani, Frolov, and Titov}{Marcheggiani
  et~al.}{2017}]{marcheggiani-2017}
Marcheggiani, D., A.~Frolov, and I.~Titov (2017, August).
\newblock A simple and accurate syntax-agnostic neural model for
  dependency-based semantic role labeling.
\newblock In {\em Proceedings of the 21st Conference on Computational Natural
  Language Learning (CoNLL 2017)}, Vancouver, Canada, pp.\  411--420.
  Association for Computational Linguistics.

\bibitem[\protect\citeauthoryear{Mikolov, Sutskever, Chen, Corrado, and
  Dean}{Mikolov et~al.}{2013}]{mikolov2013distributed}
Mikolov, T., I.~Sutskever, K.~Chen, G.~S. Corrado, and J.~Dean (2013).
\newblock Distributed representations of words and phrases and their
  compositionality.
\newblock In {\em Advances in neural information processing systems}, pp.\
  3111--3119.

\bibitem[\protect\citeauthoryear{Mulcaire, Swayamdipta, and Smith}{Mulcaire
  et~al.}{2018}]{P18-2106}
Mulcaire, P., S.~Swayamdipta, and N.~A. Smith (2018).
\newblock Polyglot semantic role labeling.
\newblock In {\em Proceedings of the 56th Annual Meeting of the Association for
  Computational Linguistics (Volume 2: Short Papers)}, pp.\  667--672.
  Association for Computational Linguistics.

\bibitem[\protect\citeauthoryear{Nair and Hinton}{Nair and
  Hinton}{2010}]{nair2010rectified}
Nair, V. and G.~E. Hinton (2010).
\newblock Rectified linear units improve restricted boltzmann machines.
\newblock In {\em Proceedings of the 27th international conference on machine
  learning (ICML-10)}, pp.\  807--814.

\bibitem[\protect\citeauthoryear{Neubig, Dyer, Goldberg, Matthews, Ammar,
  Anastasopoulos, Ballesteros, Chiang, Clothiaux, Cohn, et~al.}{Neubig
  et~al.}{2017}]{neubig2017dynet}
Neubig, G., C.~Dyer, Y.~Goldberg, A.~Matthews, W.~Ammar, A.~Anastasopoulos,
  M.~Ballesteros, D.~Chiang, D.~Clothiaux, T.~Cohn, et~al. (2017).
\newblock Dynet: The dynamic neural network toolkit.
\newblock {\em arXiv preprint arXiv:1701.03980\/}.

\bibitem[\protect\citeauthoryear{Nivre, Agi{\'c}, Ahrenberg, Aranzabe, Asahara,
  et~al.}{Nivre et~al.}{2017}]{universal_2}
Nivre, J., {\v Z}.~Agi{\'c}, L.~Ahrenberg, M.~J. Aranzabe, M.~Asahara, et~al.
  (2017).
\newblock {Universal Dependencies} 2.
\newblock {LINDAT}/{CLARIN} digital library at Institute of Formal and Applied
  Linguistics, Charles University in Prague.

\bibitem[\protect\citeauthoryear{Och and Ney}{Och and Ney}{2003}]{och2000giza}
Och, F.~J. and H.~Ney (2003).
\newblock A systematic comparison of various statistical alignment models.
\newblock {\em Computational Linguistics\/}~{\em 29\/}(1), 19--51.

\bibitem[\protect\citeauthoryear{Pad{\'o} and Lapata}{Pad{\'o} and
  Lapata}{2005}]{pado_projection}
Pad{\'o}, S. and M.~Lapata (2005).
\newblock Cross-linguistic projection of role-semantic information.
\newblock In {\em Proceedings of Human Language Technology Conference and
  Conference on Empirical Methods in Natural Language Processing}.

\bibitem[\protect\citeauthoryear{Pad{\'o} and Lapata}{Pad{\'o} and
  Lapata}{2009}]{pado2009cross}
Pad{\'o}, S. and M.~Lapata (2009).
\newblock Cross-lingual annotation projection for semantic roles.
\newblock {\em Journal of Artificial Intelligence Research\/}~{\em 36\/}(1),
  307--340.

\bibitem[\protect\citeauthoryear{Palmer, Gildea, and Kingsbury}{Palmer
  et~al.}{2005}]{palmer_propbank}
Palmer, M., D.~Gildea, and P.~Kingsbury (2005).
\newblock The proposition bank: An annotated corpus of semantic roles.
\newblock {\em Computational Linguistics, Volume 31, Number 1, March 2005\/}.

\bibitem[\protect\citeauthoryear{Petrov, Das, and McDonald}{Petrov
  et~al.}{2011}]{petrov2011universal}
Petrov, S., D.~Das, and R.~McDonald (2011).
\newblock A universal part-of-speech tagset.
\newblock {\em arXiv preprint arXiv:1104.2086\/}.

\bibitem[\protect\citeauthoryear{Rasooli, Lippincott, Habash, and
  Rambow}{Rasooli et~al.}{2014}]{rasooli-EtAl:2014:P14-1}
Rasooli, M.~S., T.~Lippincott, N.~Habash, and O.~Rambow (2014, June).
\newblock Unsupervised morphology-based vocabulary expansion.
\newblock In {\em Proceedings of the 52nd Annual Meeting of the Association for
  Computational Linguistics (Volume 1: Long Papers)}, Baltimore, Maryland, pp.\
   1349--1359. Association for Computational Linguistics.

\bibitem[\protect\citeauthoryear{Roth and Lapata}{Roth and
  Lapata}{2016}]{roth_sup}
Roth, M. and M.~Lapata (2016).
\newblock Neural semantic role labeling with dependency path embeddings.
\newblock In {\em Proceedings of the 54th Annual Meeting of the Association for
  Computational Linguistics (Volume 1: Long Papers)}, pp.\  1192--1202.
  Association for Computational Linguistics.

\bibitem[\protect\citeauthoryear{Straka and Strakov\'{a}}{Straka and
  Strakov\'{a}}{2017}]{udpipe:2017}
Straka, M. and J.~Strakov\'{a} (2017, August).
\newblock Tokenizing, pos tagging, lemmatizing and parsing ud 2.0 with udpipe.
\newblock In {\em Proceedings of the CoNLL 2017 Shared Task: Multilingual
  Parsing from Raw Text to Universal Dependencies}, Vancouver, Canada, pp.\
  88--99. Association for Computational Linguistics.

\bibitem[\protect\citeauthoryear{van~der Plas, Apidianaki, and Chen}{van~der
  Plas et~al.}{2014}]{vanderplas_cross_global}
van~der Plas, L., M.~Apidianaki, and C.~Chen (2014).
\newblock Global methods for cross-lingual semantic role and predicate
  labelling.
\newblock In {\em Proceedings of COLING 2014, the 25th International Conference
  on Computational Linguistics: Technical Papers}, pp.\  1279--1290. Dublin
  City University and Association for Computational Linguistics.

\bibitem[\protect\citeauthoryear{van~der Plas, Merlo, and Henderson}{van~der
  Plas et~al.}{2011}]{vanderplas_cross}
van~der Plas, L., P.~Merlo, and J.~Henderson (2011).
\newblock Scaling up automatic cross-lingual semantic role annotation.
\newblock In {\em Proceedings of the 49th Annual Meeting of the Association for
  Computational Linguistics: Human Language Technologies}, pp.\  299--304.
  Association for Computational Linguistics.

\bibitem[\protect\citeauthoryear{Virpioja, Smit, Gr{\"o}nroos, Kurimo,
  et~al.}{Virpioja et~al.}{2013}]{virpioja2013morfessor}
Virpioja, S., P.~Smit, S.-A. Gr{\"o}nroos, M.~Kurimo, et~al. (2013).
\newblock Morfessor 2.0: Python implementation and extensions for morfessor
  baseline.
\newblock Technical report, Aalto University.

\bibitem[\protect\citeauthoryear{Wang, Akbik, chiticariu, Li, Xia, and Xu}{Wang
  et~al.}{2017}]{wang-EtAl:2017:EMNLP20175}
Wang, C., A.~Akbik, l.~chiticariu, Y.~Li, F.~Xia, and A.~Xu (2017, September).
\newblock Crowd-in-the-loop: A hybrid approach for annotating semantic roles.
\newblock In {\em Proceedings of the 2017 Conference on Empirical Methods in
  Natural Language Processing}, Copenhagen, Denmark, pp.\  1913--1922.
  Association for Computational Linguistics.

\bibitem[\protect\citeauthoryear{Zaghouani, Diab, Mansouri, Pradhan, and
  Palmer}{Zaghouani et~al.}{2010}]{zaghouani_arabic}
Zaghouani, W., M.~Diab, A.~Mansouri, S.~Pradhan, and M.~Palmer (2010, July).
\newblock The revised arabic propbank.
\newblock In {\em Proceedings of the Fourth Linguistic Annotation Workshop},
  Uppsala, Sweden, pp.\  222--226. Association for Computational Linguistics.

\bibitem[\protect\citeauthoryear{Zhou and Xu}{Zhou and Xu}{2015}]{P15-1109}
Zhou, J. and W.~Xu (2015).
\newblock End-to-end learning of semantic role labeling using recurrent neural
  networks.
\newblock In {\em Proceedings of the 53rd Annual Meeting of the Association for
  Computational Linguistics and the 7th International Joint Conference on
  Natural Language Processing (Volume 1: Long Papers)}, pp.\  1127--1137.
  Association for Computational Linguistics.

\end{thebibliography}

\end{document}